\documentclass[preprint,12pt]{elsarticle}

\usepackage{lineno,hyperref}
\modulolinenumbers[5]

\graphicspath{ {./fig/} }
\usepackage[a4paper, total={6in, 9in}]{geometry}
\usepackage{latexsym}
\usepackage{graphicx}
\usepackage{amsfonts,amssymb,amsmath}
\usepackage{caption}
\usepackage{booktabs}
\usepackage{textcomp}
\usepackage{gensymb}
\usepackage{bm}
\usepackage{url}
\usepackage{multirow}
\usepackage{multicol}
\usepackage{lineno}
\usepackage{setspace}
\usepackage{xcolor}

\usepackage[ruled,vlined]{algorithm2e}
\usepackage{etoolbox}
\newcommand{\algorithmicdoinparallel}{\textbf{do in parallel}}
\makeatletter
\AtBeginEnvironment{algorithm2e}{%
  \newcommand{\FORALLP}[2][default]{\ALC@it\algorithmicforall\ #2\ %
    \algorithmicdoinparallel\ALC@com{#1}\begin{ALC@for}}%
}
\makeatother

\journal{Journal of \LaTeX\ Templates}

\journal{Internet of Things}

\begin{document}

\begin{frontmatter}

\title{Federated Learning with Hyperparameter-based Clustering for Electrical Load Forecasting}

 \author[ua]{Nastaran Gholizadeh}
\author[ua,uhk]{Petr Musilek\corref{cor1}}

\address[ua]{Electrical and Computer Engineering, University of Alberta, Edmonton, Alberta, Canada}
\address[uhk]{Applied Cybernetics, University of Hradec Kr\'{a}lov\'{e}, Hradec Kr\'{a}lov\'{e}, Czech Republic}
 \cortext[cor1]{Corresponding author}


\begin{abstract}
Electrical load prediction has become an integral part of power system operation. Deep learning models have found popularity for this purpose. However, to achieve a desired prediction accuracy, they require huge amounts of data for training. Sharing electricity consumption data of individual households for load prediction may compromise user privacy and can be expensive in terms of communication resources. Therefore, edge computing methods, such as federated learning, are gaining more importance for this purpose. These methods can take advantage of the data without centrally storing it. This paper evaluates the performance of federated learning for short-term forecasting of individual house loads as well as the aggregate load. It discusses the advantages and disadvantages of this method by comparing it to centralized and local learning schemes. Moreover, a new client clustering method is proposed to reduce the convergence time of federated learning. The results show that federated learning has a good performance with a minimum root mean squared error (RMSE) of 0.117kWh for individual load forecasting.
\end{abstract}


\end{frontmatter}


\section{Introduction}
Short-term load forecasting is a fundamental part of resource scheduling task in power systems~\cite{Lukumon}. Some recent studies focus on short-term load forecasting of individual buildings for decentralized monitoring and control of power systems~\cite{SKOMSKI2020,KIM2019}. This has become important due to the distributed placement of intermittent renewable energy resources. However, load forecasting of individual households is a challenging task since the house load profile is highly dependent on the stochastic behavior of its residents~\cite{Yaxing}.

Many algorithms have been proposed for short-term load forecasting in literature. They include Auto Regressive Integrated Moving Average (ARIMA)~\cite{VANDERMEER} model, Markov-chain Mixture (MCM) distribution model~\cite{MUNKHAMMAR2021}, Quantile Regression (QR)~\cite{Taieb,forecast101} method, Support Vector Regression (SVR)~\cite{Yaslan}, and deep learning models. Especially, certain deep learning models, such as Long Short Term Memory (LSTM) networks~\cite{Woohyun, Memarzadeh}, hybrid Convolutional Neural Network (CNN) and LSTM networks~\cite{Xifeng}, and some other types of hybrid recurrent neural networks and CNNs~\cite{Zulfiqar}, have shown good performance for short-term load forecasting. These methods require high volumes of historical data for training and they store the data centrally. Transferring and centrally storing the historical energy consumption data of individual buildings is expensive in terms of communication resources and it can compromise electricity user's privacy. Moreover, with the proliferation of advanced metering infrastructure such as smart meters~\cite{Godoy,Vilaplana}, it is infeasible to centrally store and manage the massive amounts of data produced by these devices. Edge computing and distributed learning methods are promising solutions to overcome these issues by eliminating the need for central data storage and processing. These methods also reduce the computational burden of training deep learning models by dividing the processing task among multiple edge devices.

Federated learning, introduced by Google in 2017~\cite{Brendan}, is a type of distributed machine learning mechanism that allows training a global model by cooperation of edge devices and without sharing the training data. In this method, the training is performed by the edge devices and the resultant weights are shared with the central server to perform weight updates. The updated weights are then sent back to the edge devices for another round of training. This iteration continues until model accuracy is high enough~\cite{Gholizadeh}. Since no raw data is exchanged between the edge devices and each edge device carries out a part of the training task, federated learning preserves data privacy and has higher scalability compared to centralized learning methods \cite{MESSAOUD2020}.

Due to the numerous benefits it offers, federated learning is quickly expanding to various research areas such as healthcare~\cite{Brisimi}, communications~\cite{Subramanya}, language modeling~\cite{FedMed}, transportation~\cite{Railway}, and many others . A few studies have also used federated learning for short-term electrical load forecasting. To this end, Savi and Olivadese~\cite{Savi} developed a federated LSTM model to forecast electrcity demand of individual houses and compared it to the centralized model. To enhance the forecasting accuracy, customers were clustered based on socioeconomic affinities and consumption similarities such as average, total, and lowest energy consumption. However, these indices may not properly reflect the consumption patterns of consumers. Therefore, clustering might not enhance the model performance. Taïk and Cherkaoui~\cite{Cherkaoui} used federated LSTM for the same purpose but without customer clustering. The results showed that the performance of federated learning for load forecasting was not good even after a few rounds of personalization. Personalization is the process of taking the global model and running a few local training rounds on it to increase prediction accuracy. Tun et al.~\cite{Choong} used bidirectional LSTM with OPTICS clustering method to group consumers based on house type, region, facing direction, number of rental units, and heating type. However, these attributes cannot reflect consumer behavior and load shape properly as they remain constant even if the residents of a dwelling change. Horizontal and vertical federated learning were applied by Liu et al.~\cite{Haizhou} for power consumption prediction without customer clustering. The main goal was to demonstrate how much federated learning can preserve privacy. 

Although there have been a few studies published in the area of federated learning for energy forecasting, its performance for forecasting of individual buildings' energy consumption and aggregate demands has not been thoroughly studied yet. This paper assesses the performance of federated learning for predicting both individual house electrical loads as well as aggregate electricity demand by comparing them to centralized and local forecasting schemes. Moreover, a new customer clustering method is introduced which does not require sharing any confidential customer data. The main contributions of this paper are as follows:

\begin{itemize}
\item {Conducting a thorough study of federated learning performance compared to centralized and local forecasting for predicting single and aggregate electrical demands;}
\item {Proposing a new consumer clustering technique that better reflects consumer consumption pattern and does not require sharing confidential consumer data;}
\item {Implementing an algorithm into the federated learning that detects and removes consumers that deter the global model.}
\end{itemize}

The remainder of this paper is organized as follows. Section \ref{sys_model} describes the federated learning framework used in this study. Simulation setup, used data, and results are presented in Section \ref{results}. Finally, the concluding remarks are drawn in Section \ref{Conclusion}.

\section{System Model}\label{sys_model}
This section describes the system model for applying federated learning and edge computing to forecast the the electrical demand of single houses as well as the aggregate demands.

\subsection{Federated Learning Model}
\begin{figure}[h]
	\begin{center}
		\includegraphics[width=0.8\columnwidth]{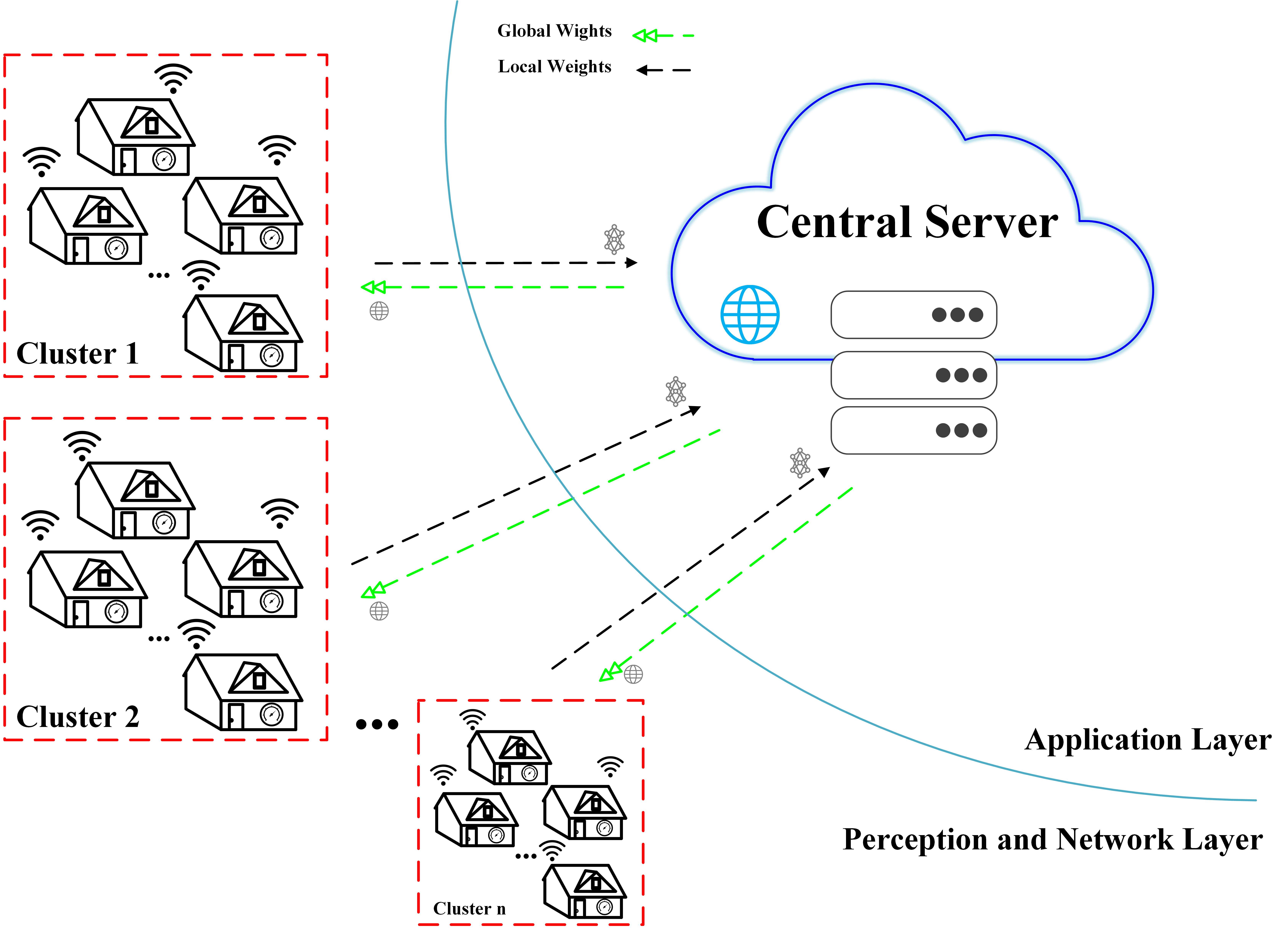}\\
		\caption{Federated learning scheme}\label{f1}
	\end{center}
\end{figure}

The architecture of the system used in this study is illustrated in Figure \ref{f1}. The system is divided into two subsystems: central server (cloud) and clients (individual houses). Each client contains a smart meter device that measures the electricity consumption and communicates with the central server. Direct smart meter to cloud communication is feasible using the current technology \cite{PAU2018,EMnify}. In addition, it is assumed that these smart meters contain processing units to train simple neural networks using the historical energy consumption data stored within them. In federated learning, the central server sends the initial weight values to clients to start the training. After one or more steps of training by each client, the obtained weights are sent back to the central server to perform wight update after aggregating weights from all clients. The central server uses FedAvg for this purpose. FedAvg method basically performs a weighted averaging on the weights obtained from all clients. This can be shown as \begin{equation}
\bold w_{k+1}=\sum_{c=1}^C \frac{n_c}{n}\nabla \bold w_c,
\label{e1}
\end{equation}
where $\bold w_{k+1}$ is the updated weights sent by the central server in the $(k+1)$th round of communication and $\bold w_c$ is the weights sent by the client $c$ to the central server. Parameter $n$ is the total number of data points used for training the global model and $n_c$ is the total number of data points used by client $c$ for training. 

The central server sends the updated weights back to the clients for another round of training and this iteration between clients and the central server continues until desired training accuracy is achieved. As can be understood from this process, FedAvg attempts to fit a general model to the clients' data and each client contributes to the global model based on the amount of data it used for training. The process of federated learning is shown in Algorithm \ref{A1} in more details.

\setlength{\textfloatsep}{0.1cm}
\begin{algorithm}[h!]
\footnotesize
\nl\textbf{Input:}~~ Hyperparameter data of clients \par
\nl\KwOut{Cluster-specific trained models} \par
$\triangleright$~~Initialization \par
\nl{\For{each cluster $k$ with $k = 1, 2, ..., N$ {\normalfont\textbf{in parallel}}}{
\nl Initialize weights $w_c$ \par
\nl Update local models of clients using these weights \par
$\triangleright$~~Communication loop \par
\nl \For{Communication round $t = 1, 2, ..., T$}{
\nl \For{each client $c$ with $c = 1, 2, ..., C$ {\normalfont\textbf{in parallel}}}{
\nl Train and update the local model using local data \par
\nl Send updated weights to the central server} \par
$\triangleright$~~Detecting and removing clients that deter the model \par
\nl \If{$loss_{c, T} > 1.6 \times loss_{c, T-20}$}{
\nl Remove client $c$} \par
$\triangleright$~~Federated averaging \par
\nl Aggregate weights and update global model using FedAvg method in Eq. \ref{e1} }}}
\caption{{\bf Federated learning with clustered aggregation} \label{A1}}
\end{algorithm}

As it can be seen, the three-layer IoT architecture is used here. The perception and network layers consist of smart meters that measure energy consumption and send it to the central server. The application layer includes a central server that performs computations on the data received from clients. The smart meters use MQTT paired with TCP/IP internet protocol suite for sending data to the central server. Each smart meter pushes weight data to the central server and polls weight updates from the server. Each smart meter is specified by its respective client throughout this paper. Details regarding IoT communication and security requirements are out of the scope of this paper.

\subsection{LSTM Model}

\begin{figure}[h!]
	\begin{center}
		\includegraphics[scale=0.5]{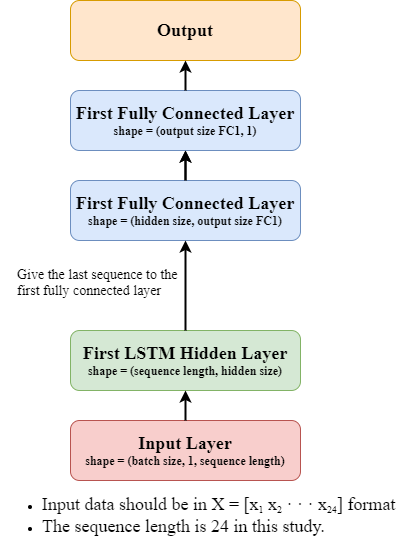}\\
		\caption{Structure of the LSTM network used in this study}\label{lstm}
	\end{center}
\end{figure}

LSTM networks are a special type of recurrent neural networks (RNN) that are able to learn long-term temporal correlations in data sequences and were developed to solve the vanishing gradient problem in original RNNs. They have a forget gate that decides what information to keep or throw away during the learning phase. Recent research suggests that LSTM and its variants have the best performance for time-series forecasting~\cite{Muzaffar, Kharlova20}. Therefore, this paper uses LSTM networks for the training phase inside the clients. In this study, the time series data is sampled into sliding windows with look-back size of 24 hours. These sliding windows are the inputs to the LSTM model and the output is the next-hour electricity load. The structure of the LSTM network used in this study is shown in Figure \ref{lstm}. This structure is similar to the model developed in~\cite{9097842}. The proposed network has one hidden LSTM layer and two fully connected layers.

It should be noted that the local training using historical data and LSTM networks is performed inside each client while the global update is performed by the central server after aggregating the neural network weights obtained by the clients. The overall framework is called federated learning.

\subsection{Customer Clustering}
Since in the FedAvg method, the obtained local weights are averaged by the central server, it is important for clients' load profiles to have similar patterns. Otherwise, the heterogeneity in data may lead to misconvergence of the global model \cite{misconvergence}. For example, a client with a very different and unique load profile might affect the global model weights to fit its own data. This will cause the global model to diverge from fitting other clients' data and, consequently, the error of the obtained model will remain high. To prevent this issue, the clients must be grouped based on their similarities. However, since the global model has no access to client data in federated learning, customer clustering is a challenging task.

Some research works have used clustering based on socioeconomic affinities, metrics such as average, total, and lowest energy consumption, and some other data such as house type, region, facing direction, and number of rental units. However, none of these metrics can reveal consumption patterns properly. A method was proposed by Xu et al.~\cite{Haohang} to cluster clients based on their obtained weights after the first communication round with the central server. This method might not be accurate either since it takes more training rounds for client weights to converge. In fact, the initial weights adjusted by the central server are the same for all clients at the beginning of the training and might not significantly change after only a few training steps.

This paper proposes a new clustering method that is able to reveal the similarity between load profiles of various clients. Since all clients are equipped with processing units needed for training, they can also perform hyperparameter tuning for a simple neural network using their data. In this method, the central server asks clients to perform hyperparameter tuning for their training network or another simpler neural network. These hyperparameters can include number of neurons in each hidden layer, number of epochs for training, learning rates, etc., and the tuning can be easily performed using GridSearchCV method. Each client updates its hyperparameter tuning results from time to time and shares this non-sensitive data with the central server. The central server clusters the clients based on their hyperparameters when performing federated learning. This way, the clients are more likely to be grouped based on their data similarities as their hyperparameters are selected based on the historical data stored in each client. In previous studies, the neural network structure for all of the clusters was the same. In the proposed method, the central server fits a specific neural network model to each cluster based on the obtained hyperparameters for that cluster. \textcolor{black}{The number of clusters is determined by the central server based on various factors such as accuracy, communication/computation cost, or trade-off between communication/computation cost and accuracy. In addition, the proposed method has the flexibility to adapt to changes in number of clients or their consumption patterns.}

In this paper, an algorithm is implemented within the federated learning process that detects and removes clients that deter the global model. For this purpose, the training loss of clients in each round is assessed and compared to previous rounds to check whether it is getting smaller or not. In this study, the clients whose loss is 60\% worse compared to 20 rounds ago are removed from the federated learning process as shown by lines 10-11 in Algorithm \ref{A1}.

\section{Simulation and Results}\label{results}
This section describes the dataset and the parameters of the model used in this study. Subsequently, it presents, discusses, and compares the obtained results.

\subsection{Simulation Data}
This study was conducted using data from 75 locations within the City of Edmonton. The data was provided by EPCOR distribution company under the APIC-alliance project and is not available online. \textcolor{black}{The data from residential buildings is used in this study and clustering is used to group them based on similarity in their consumption behavior.} The proposed LSTM network consists of one hidden layer. The output of this layer is given to two fully connected hidden layers. Each client performs hyperparameter tuning for this network using its local data to find the best number of neurons for the first and second fully connected layers as well as the number of training epochs. The clients share this data with the central server and the server groups the clients into five clusters using the k-means clustering method. The hyperparameters with the highest number of occurrences within each cluster are chosen for the general structure of that cluster. Inside each client, 75\% of data is used for training and 25\% for testing. The programming was performed in Python using PyTorch library, on a 3.6GHz processor with 48GB RAM.

\begin{figure}[h]
	\begin{center}
		\includegraphics[width=0.6\columnwidth]{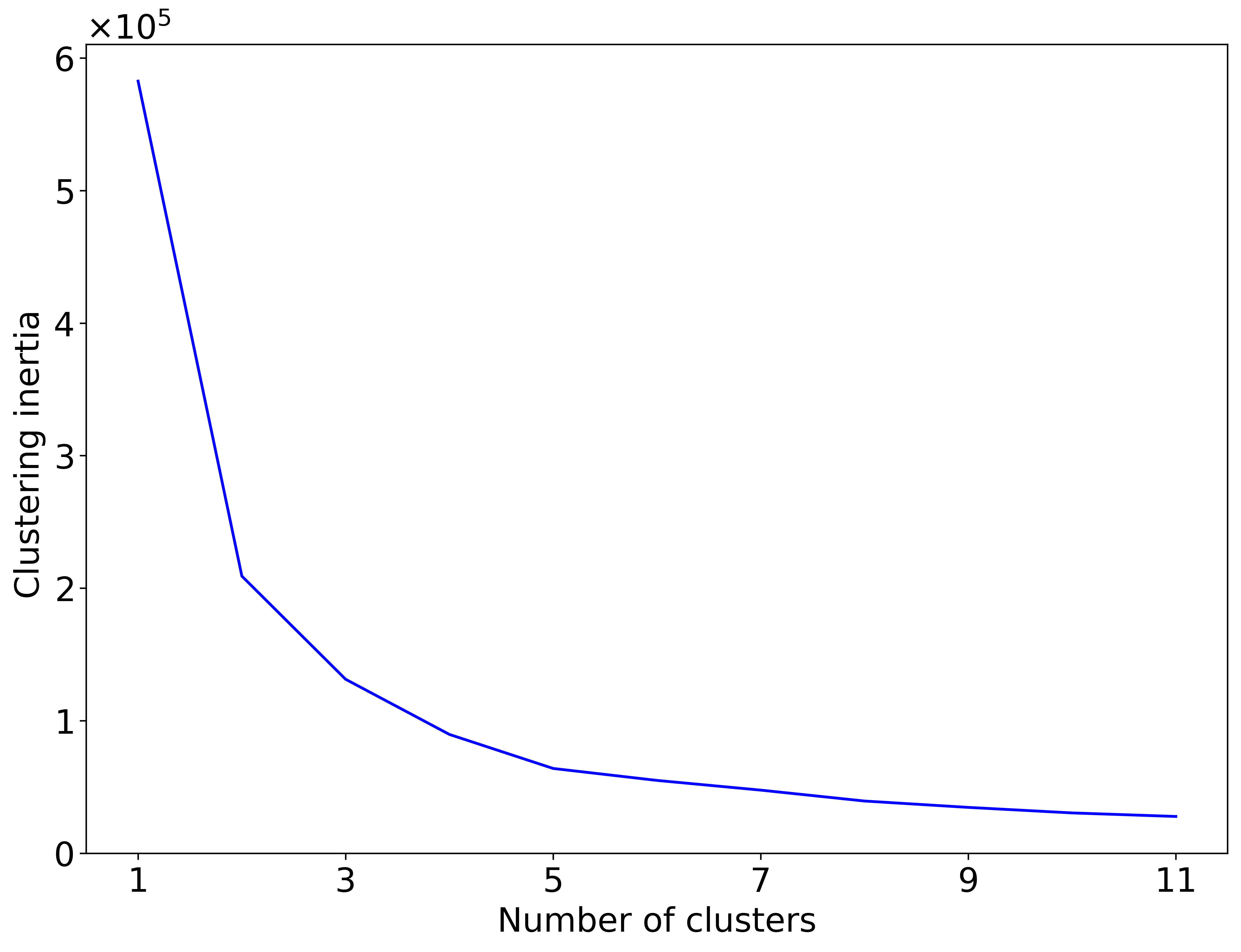}\\
		\caption{\textcolor{black}{Clustering inertia for various numbers of clusters using k-means clustering}}\label{cluster}
	\end{center}
\end{figure}

The hyperparameters for each cluster are given in Table \ref{t1}. \textcolor{black}{To determine the best number of customer clusters in this study, k-means clustering was repeated for various number of clusters and the obtained inertia (sum of squared distances of samples to their closest cluster center) was plotted against the number of clusters as shown in Figure \ref{cluster}. The five-cluster model was selected since after this point in the plot, there is no significant decrease in inertia. This choice of the lowest acceptable number of clusters optimizes the communication/computation cost of distributed learning.}

\begin{table}[h!]
\centering
\caption{Hyperparameters for clusters}
\label{t1}
\resizebox{0.75\columnwidth}{!} {
\begin{tabular}{|c|c|c|c|c|}
\cline{1-5}
 Cluster\#  & First layer neurons & Second layer neurons & Epochs & Number of clients\\
\hline
Cluster 1   & 44 & 127 & 148 & 11\\
\hline
Cluster 2   & 68 & 198 & 247 & 9\\
\hline
Cluster 3 & 56 & 198 & 103 & 15\\
\hline
Cluster 4 & 32 & 85 & 291 & 17\\
\hline
Cluster 5 & 32 & 127 & 103 & 23\\
\hline
\end{tabular}}
\end{table}

Mean squared error (MSE) is used as the loss function in this study\begin{equation}
MSE=\frac{\sum_{i=1}^N (y_i-\hat{y_i})^2}{N},
\label{e2} 
\end{equation}
where $y_i$ is the actual value, $\hat{y_i}$ is the predicted value, and $N$ is the total number of predicted values.

\subsection{Simulation Results}
To evaluate the performance of federated learning for individual load prediction (ILP) and aggregate load prediction (ALP), the results are compared with local and centralized forecasts. Local and centralized learning are defined as follows:

\begin{itemize}
\item {Centralized learning: data is gathered from all clients and a single central server performs training and forecasting.}
\item {Local learning: data stays within each client and each client performs training and forecasting for its local data.}
\end{itemize}

Root mean squared error (RMSE) is selected as the evaluation criteria in this study as it has the same unit as the actual load and can be easily compared to it. Mean absolute percentage error (MAPE) is not preferred for load forecast problems since load values, especially in ILP, are usually small. This can result in large MAPE values which do not correctly reflect the performance of the forecasting algorithms~\cite{Davydenko2014}.

Table \ref{t2} presents the obtained RMSE results for cluster 1. The best obtained results are typeset in bold. The forecasting RMSE can be compared with the maximum and minimum energy consumption given in the same table to get a sense of prediction accuracy. As can be seen, the performance of federated learning for ILP is comparable and, in some cases, even better than local learning. \textcolor{black}{It was also observed that,} depending on the initial random weights, the performance of local learning can vary to a large extent and easily get trapped in a local minimum. However, the global updates in federated learning can circumvent \textcolor{black}{this problem}. \textcolor{black}{To mitigate the local minimum issue in local learning, three different optimization algorithms (stochastic gradient descent, RMSprop, and Adam) were tested, and Adam was selected due to its better performance. In addition, the learning rate, number of neurons in each layer, and number of iterations were selected to minimize the mean squared error. The parameters were consistent in federated, local, and central learning to allow fair comparison. The mean RMSE in the same table reveals that, on average, local learning performs best for ILP.} For the ALP, centralized learning performs better than local and federated learning. It should be noted that, in local learning for ALP, each client performs the forecast individually and the results are summed up to get the aggregate demand.

\begin{table}[h!]
\centering
\caption{Prediction results for cluster 1 (units in kWh)}
\label{t2}
\resizebox{0.9\columnwidth}{!} {
\begin{tabular}{|c|c|c|c|c|c|}
\cline{4-6}
\multicolumn{3}{c|}{} & Federated learning & Centralized learning & Local learning\\
\hline
\multirow{4}{*}{Test Dataset}   & \multirow{3}{*}{RMSE (ILP)} & Min & 0.187 & 1 & \textcolor{black}{\textbf{0.159}} \\
\cline{3-6}
   &  & Max & 0.597 & 1.115 &  \textcolor{black}{\textbf{0.504}}\\
\cline{3-6}
   &  & Mean & 0.418 & 1.062 &  \textcolor{black}{\textbf{0.334}}\\
\cline{2-6}
   & RMSE (ALP) & - & 2.205 & \textbf{1.152} &  \textcolor{black}{1.176}\\
\hline
\multirow{4}{*}{Train Dataset}   & \multirow{3}{*}{RMSE (ILP)} & Min & 0.205 & 1.005 &  \textcolor{black}{\textbf{0.139}}\\
\cline{3-6}
   &  & Max & \textbf{0.396} & 0.63 &  \textcolor{black}{0.512}\\
\cline{3-6}
   &  & Mean & 0.4 & 1.06 &  \textcolor{black}{\textbf{0.306}}\\
\cline{2-6}
   & RMSE (ALP) & - & 2.022 & 1.022 &  \textcolor{black}{\textbf{0.981}}\\
\hline
\multicolumn{2}{|c|}{Max client energy consumption} & 6.378 & \multicolumn{2}{|c|}{Mean client energy consumption} & 0.546\\
\hline
\multicolumn{2}{|c|}{Max aggregate energy consumption} & 12.832 & \multicolumn{2}{|c|}{Mean aggregate energy consumption} & 4.37\\
\hline
\end{tabular}}
\vspace{0.5cm}
\end{table}

Prediction results for cluster 2 are given in Table \ref{t3}. Similar to cluster 1, federated learning performs better than centralized learning and is comparable to local learning for ILP. \textcolor{black}{Centralized learning has the best performance for ALP and local learning has similar results.}\\

\begin{table}[h!]
\centering
\caption{Prediction results for cluster 2 (units in kWh)}
\label{t3}
\resizebox{0.9\columnwidth}{!} {
\begin{tabular}{|c|c|c|c|c|c|}
\cline{4-6}
\multicolumn{3}{c|}{} & Federated learning & Centralized learning & Local learning\\
\hline
\multirow{4}{*}{Test Dataset}   & \multirow{3}{*}{RMSE (ILP)} & Min & 0.308 & 2.342& \textbf{0.272}\\
\cline{3-6}
   &  & Max & \textbf{1.055} &  2.903 & \textcolor{black}{1.155}\\
\cline{3-6}
   &  & Mean & 0.607 &  2.627 & \textcolor{black}{\textbf{0.479}}\\
\cline{2-6}
   & RMSE (ALP) & - & 3.904 & \textbf{1.448} & \textcolor{black}{1.741}\\
\hline
\multirow{4}{*}{Train Dataset}   & \multirow{3}{*}{RMSE (ILP)} & Min & 0.213 & 2.124 & \textbf{0.17}\\
\cline{3-6}
   &  & Max & 1.288  & 2.903  & \textcolor{black}{\textbf{0.713}}\\
\cline{3-6}
   &  & Mean & 0.568  & 2.614 & \textcolor{black}{\textbf{0.330}}\\
\cline{2-6}
   & RMSE (ALP) & - & 4.003  & 1.272 & \textcolor{black}{\textbf{1.162}}\\
\hline
\multicolumn{2}{|c|}{Max client energy consumption} & 10.959 & \multicolumn{2}{|c|}{Mean client energy consumption} & 1.33\\
\hline
\multicolumn{2}{|c|}{Max aggregate energy consumption} & 22.399 & \multicolumn{2}{|c|}{Mean aggregate energy consumption} & 11.978\\
\hline
\end{tabular}}
\vspace{0.3 cm}
\end{table}

Tables \ref{t4}-\ref{t6} show the results obtained for clusters 3-5. The observed results show a similar trend as clusters 1 and 2 \textcolor{black}{except that, in clusters 3 and 4, local learning performs slightly better for ALP}. In general, \textcolor{black}{the best case for federated learning is cluster 5 with an average RMSE of 0.433 kWh that is 14.55\% higher than the average RMSE} using local learning. In the worst case, which occurs \textcolor{black}{for cluster 4, the average RMSE obtained using federated learning is 0.874 kWh which is 40.74\%} higher than the \textcolor{black}{mean} RMSE obtained by local learning.

\begin{table}[h!]
\centering
\caption{Prediction results for cluster 3 (units in kWh)}
\label{t4}
\resizebox{0.9\columnwidth}{!} {
\begin{tabular}{|c|c|c|c|c|c|}
\cline{4-6}
\multicolumn{3}{c|}{} & Federated learning & Centralized learning & Local learning\\
\hline
\multirow{4}{*}{Test Dataset}   & \multirow{3}{*}{RMSE (ILP)} & Min & 0.186 & 1.81 & \textbf{0.173}\\
\cline{3-6}
   &  & Max & 1.276 & 2.347 & \textcolor{black}{\textbf{1.033}}\\
\cline{3-6}
   &  & Mean & 0.627 & 2.153 & \textcolor{black}{\textbf{0.534}}\\
\cline{2-6}
   & RMSE (ALP) & - & 4.016 & 2.944 & \textcolor{black}{\textbf{2.782}}\\
\hline
\multirow{4}{*}{Train Dataset}   & \multirow{3}{*}{RMSE (ILP)} & Min & 0.189 & 1.784 & \textcolor{black}{\textbf{0.176}}\\
\cline{3-6}
   &  & Max & 1.331 & 2.334 & \textcolor{black}{\textbf{1.00}}\\
\cline{3-6}
   &  & Mean & 0.606 & 2.14 & \textcolor{black}{\textbf{0.487}}\\
\cline{2-6}
   & RMSE (ALP) & - & 3.953 & 2.620 & \textcolor{black}{\textbf{2.557}}\\
\hline
\multicolumn{2}{|c|}{Max client energy consumption} & 2.927 & \multicolumn{2}{|c|}{Mean client energy consumption} & 1.286\\
\hline
\multicolumn{2}{|c|}{Max aggregate energy consumption} & 36.93 & \multicolumn{2}{|c|}{Mean aggregate energy consumption} & 19.298\\
\hline
\end{tabular}}
\end{table}

\begin{table}[h!]
\centering
\caption{Prediction results for cluster 4 (units in kWh)}
\label{t5}
\resizebox{0.9\columnwidth}{!} {
\begin{tabular}{|c|c|c|c|c|c|}
\cline{4-6}
\multicolumn{3}{c|}{} & Federated learning & Centralized learning & Local learning\\
\hline
\multirow{4}{*}{Test Dataset}   & \multirow{3}{*}{RMSE (ILP)} & Min & 0.237 & 5.973 & \textcolor{black}{\textbf{0.130}}\\
\cline{3-6}
   &  & Max & 3.199 & 8.706 & \textcolor{black}{\textbf{1.653}}\\
\cline{3-6}
   &  & Mean & 0.874 & 8.314 & \textcolor{black}{\textbf{0.621}}\\
\cline{2-6}
   & RMSE (ALP) & - & 3.939 & 3.265 & \textcolor{black}{\textbf{2.865}}\\
\hline
\multirow{4}{*}{Train Dataset}   & \multirow{3}{*}{RMSE (ILP)} & Min & 0.268 & 6.013 & \textbf{0.119}\\
\cline{3-6}
   &  & Max & 3.337 & 8.703 & \textcolor{black}{\textbf{1.548}}\\
\cline{3-6}
   &  & Mean & 0.888 & 8.329 & \textcolor{black}{\textbf{0.555}}\\
\cline{2-6}
   & RMSE (ALP) & - & 4.235 & 3.498 & \textcolor{black}{\textbf{2.860}}\\
\hline
\multicolumn{2}{|c|}{Max client energy consumption} & 6.515 & \multicolumn{2}{|c|}{Mean client energy consumption} & 1.593\\
\hline
\multicolumn{2}{|c|}{Max aggregate energy consumption} & 50.155 & \multicolumn{2}{|c|}{Mean aggregate energy consumption} & 25.494\\
\hline
\end{tabular}}
\vspace{0.5cm}
\end{table}

\begin{table}[h!]
\centering
\caption{Prediction results for cluster 5 (units in kWh)}
\label{t6}
\resizebox{0.9\columnwidth}{!} {
\begin{tabular}{|c|c|c|c|c|c|}
\cline{4-6}
\multicolumn{3}{c|}{} & Federated learning & Centralized learning & Local learning\\
\hline
\multirow{4}{*}{Test Dataset}   & \multirow{3}{*}{RMSE (ILP)} & Min & 0.117 & 4.236 & \textbf{0.088}\\
\cline{3-6}
   &  & Max & \textbf{1.095} & 4.836 & \textcolor{black}{1.471}\\
\cline{3-6}
   &  & Mean & 0.433 & 4.676 & \textcolor{black}{\textbf{0.378}}\\
\cline{2-6}
   & RMSE (ALP) & - & 2.7 & \textbf{2.194} & \textcolor{black}{2.630}\\
\hline
\multirow{4}{*}{Train Dataset}   & \multirow{3}{*}{RMSE (ILP)} & Min & 0.13 & 4.204 & \textbf{0.097}\\
\cline{3-6}
   &  & Max & 1.088 & 4.831 & \textcolor{black}{\textbf{0.759}}\\
\cline{3-6}
   &  & Mean & 0.406 & 4.686 & \textcolor{black}{\textbf{0.338}}\\
\cline{2-6}
   & RMSE (ALP) & - & 2.429 & \textbf{0.311} & \textcolor{black}{2.068}\\
\hline
\multicolumn{2}{|c|}{Max client energy consumption} & 2.011 & \multicolumn{2}{|c|}{Mean client energy consumption} & 0.717\\
\hline
\multicolumn{2}{|c|}{Max aggregate energy consumption} & 29.631 & \multicolumn{2}{|c|}{Mean aggregate energy consumption} & 16.496\\
\hline
\end{tabular}}
\vspace{0.5cm}
\end{table}

\begin{figure}[h!]
	\begin{center}
		\includegraphics[width=0.8\columnwidth]{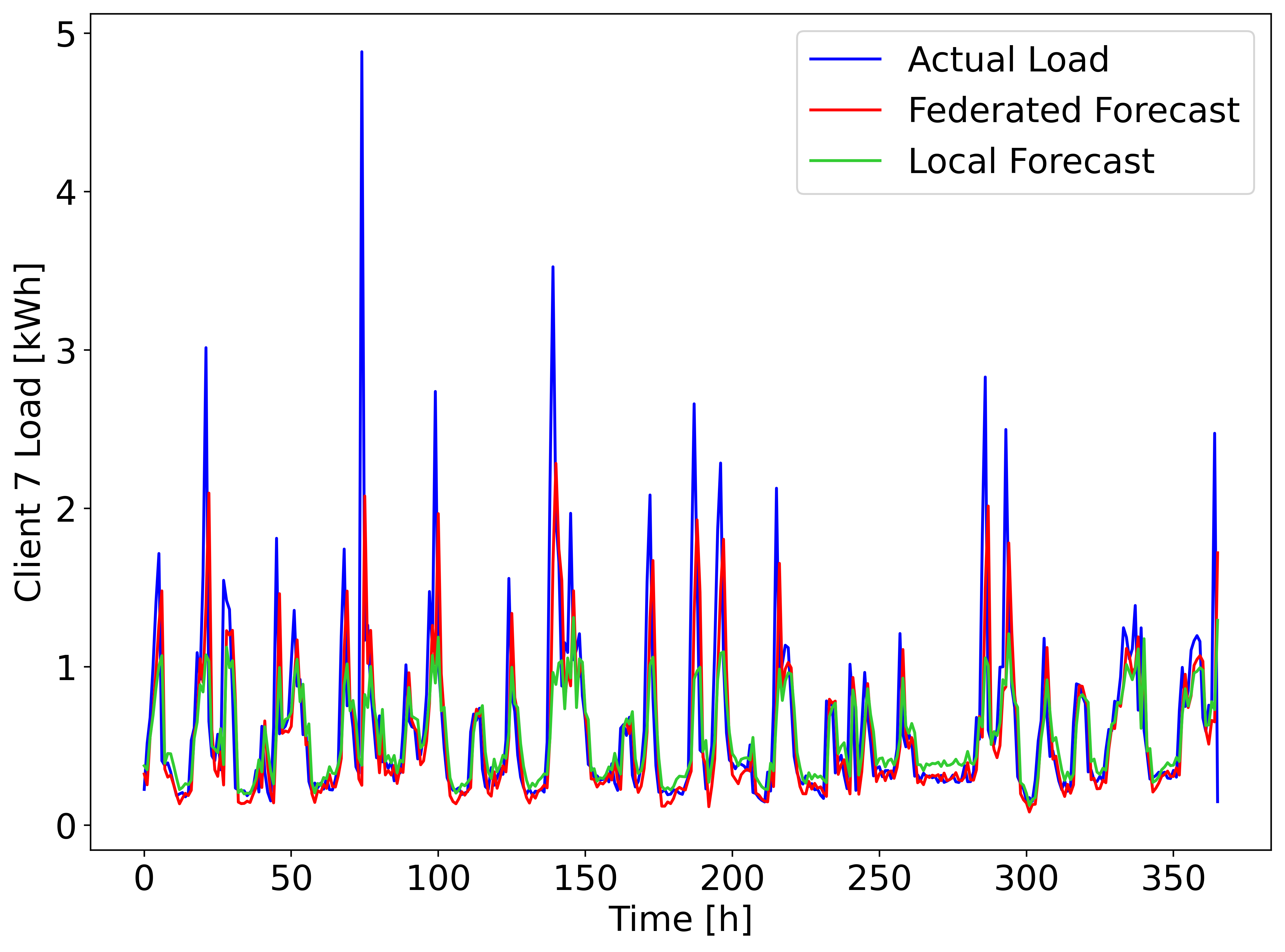}\\
		\caption{Forecast results for electrical load of client 7 inside cluster 3 using federated and local learning}\label{f2}
	\end{center}
\end{figure}

\begin{figure}[h!]
	\begin{center}
		\includegraphics[width=0.8\columnwidth]{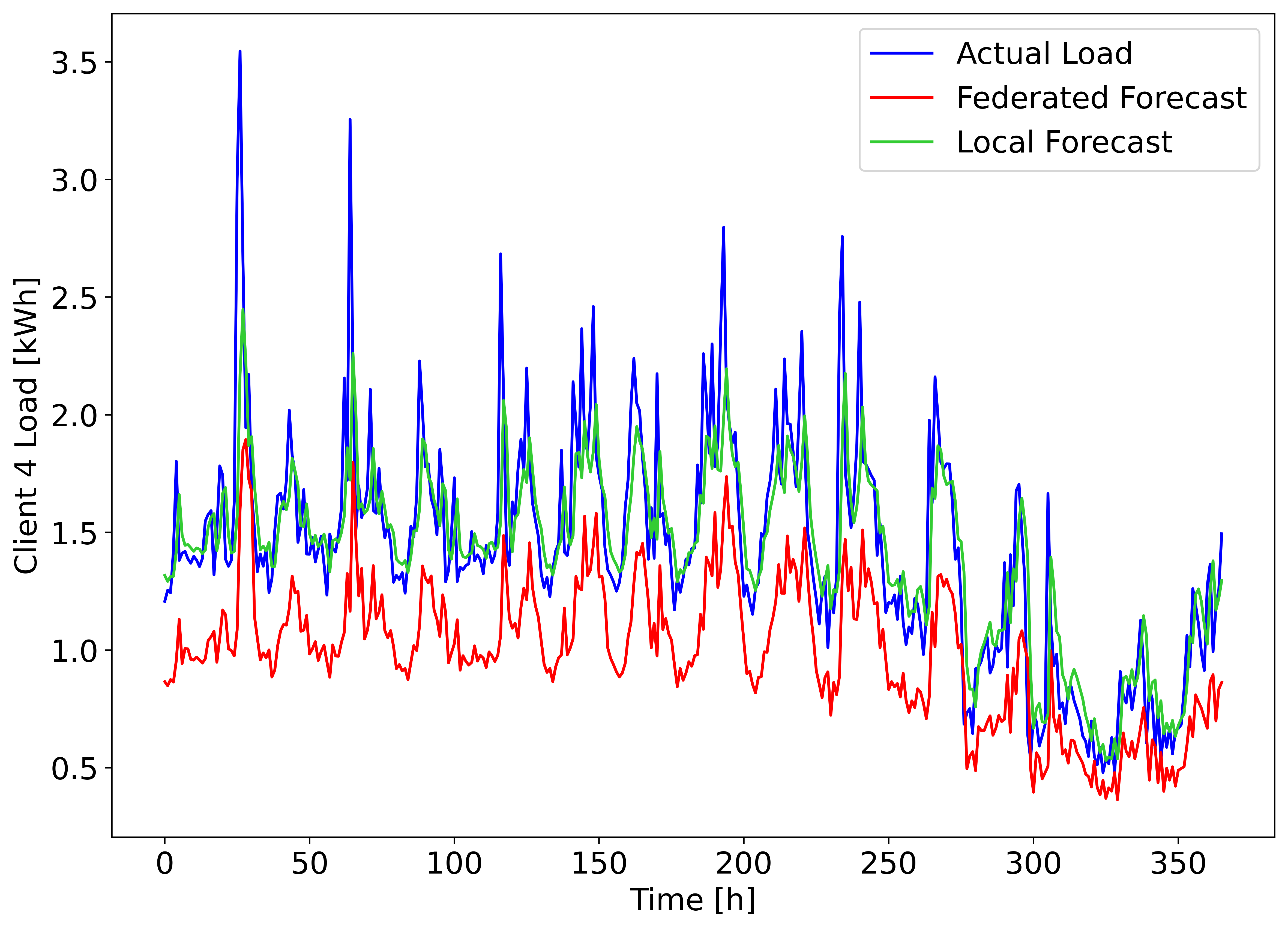}\\
		\caption{Forecast results for electrical load of client 4 inside cluster 2 using federated and local learning}\label{f3}
	\end{center}
\end{figure}

Figure \ref{f2} illustrates the federated and local forecast for ILP on the test dataset of client 7 inside cluster 3. It can be observed that, in this case, federated learning performs better than local learning. On the other hand, Figure \ref{f3} shows an example where local learning performs better than federated learning for ILP. Centralized learning does not perform well for ILP and cannot be compared to federated and local learning. Therefore, it is not presented in these figures. \textcolor{black}{It can be observed that although federated learning was trained using heterogeneous data from all customers, it still has the capability to perform well on ILP. Considering other benefits that federated learning offers, such as increased data privacy, reduced need for central data storage, and reduced computational burden due to the use of edge computing, it can be a good alternative to local and central learning for ILP.}

\begin{figure}[h!]
	\begin{center}
		\includegraphics[width=0.8\columnwidth]{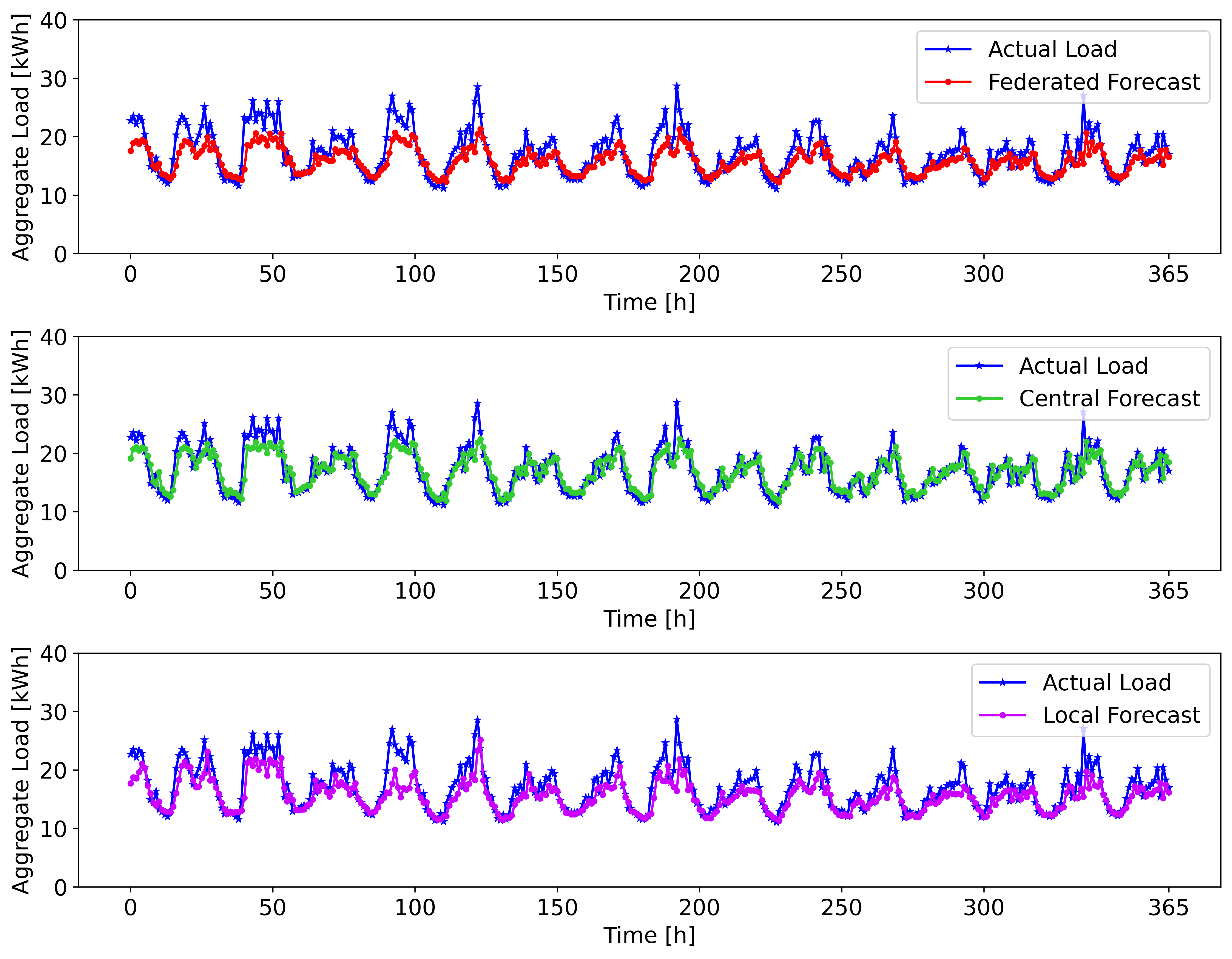}\\
		\caption{Forecast results for aggregate load of cluster 5 using federated, centralized, and local learning}\label{f4}
	\end{center}
\end{figure}

Figure \ref{f4} compares the forecast results from federated, centralized, and local learning for ALP. As seen before, the centralized forecast performs best for ALP \textcolor{black}{and is closely followed by local forecast. However, in absolute terms, federated learning's performance for ALP is not good enough. Therefore, it should only be used in cases where centralized learning and access to data is not possible.} An example of such a case is forecasting the aggregate demand of houses that are not connected to the same distribution substation while preserving privacy of users.\\

\begin{figure}[h!]
	\begin{center}
		\includegraphics[width=0.8\columnwidth]{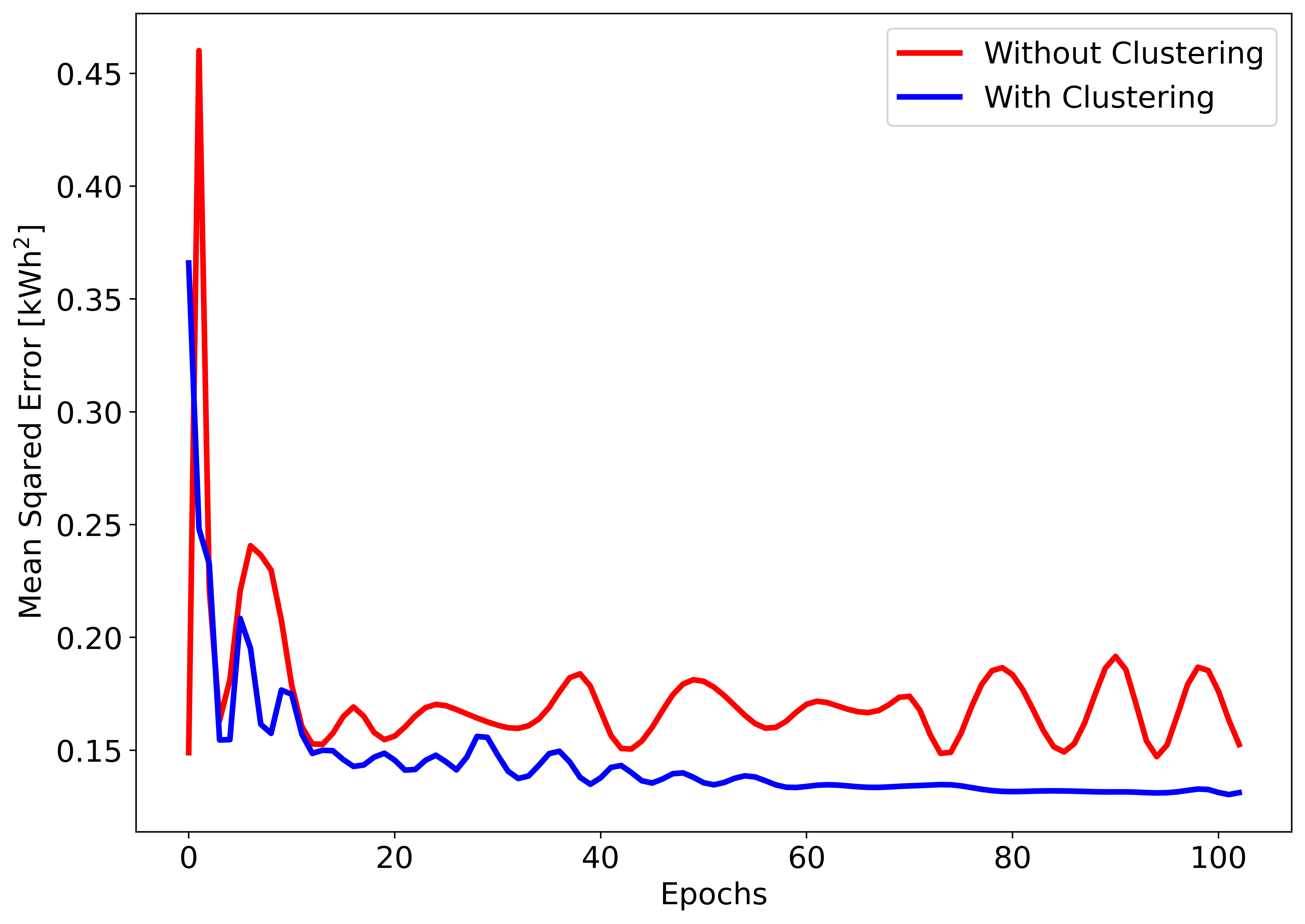}\\
		\caption{A comparison of the convergence of federated learning with and without hyperparameter-based clustering for client 7 (cluster 3)}\label{f5}
	\end{center}
\end{figure}

To show the effectiveness of the clustering method proposed in this study, MSE of federated learning with and without clustering for client 7 inside cluster 3 is illustrated in Figure \ref{f5}. It can be inferred from this figure that, the training loss without clustering takes more epochs to converge and is higher compared to training with clustering. The results were repeated and verified using data from other clients as well.

\section{Conclusion}\label{Conclusion}
Federated learning with clustered aggregation was implemented and compared to centralized and local learning for individual and aggregate load forecast problems. A new clustering method based on hyperparameter tuning was proposed and tested in this study. This new approach clusters the clients based on their data similarity while preserving their privacy. The results confirmed that the proposed clustering method reduces the convergence time of federated learning. The study was conducted by dividing the data from 76 houses into five clusters. The results showed that, \textcolor{black}{on average,} federated learning's performance was weaker than local learning for forecasting individual loads. \textcolor{black}{In the worst case, federated learning's mean RMSE within the cluster was 40.74\% higher and in the best case, it was 14.55\% higher than local learning. However, assessing the results for each house revealed that in some cases, federated learning had lower RMSE than local learning.} In addition, it is very common that local learning gets trapped in a local minimum during the training process. \textcolor{black}{Such problems never occurred during the federated learning due to the global updates.} The federated learning's performance was always better than centralized learning for individual load forecast. \textcolor{black}{In the aggregate load forecast problem, federated learning did not perform well in any of the clusters.}

In general, it can be concluded that federated learning often gives reliable results for the individual load forecast problem and it can be used in cases where access to training data is not possible or it is restricted due to privacy issues. Using federated learning for aggregate demand prediction is not recommended except in cases where access to aggregate data is not possible, as discussed in the paper. Future research should focus on optimizing the number of local training epochs by each client to synchronize the communication time with the central server and to increase the overall forecasting accuracy of federated learning.

\section{Acknowledgments}
This research was funded by the Natural Sciences and Engineering Research Council (NSERC) of Canada grant number ALLRP 549804-19 and by Alberta Electric System Operator, AltaLink, ATCO Electric, ENMAX, EPCOR Inc., and FortisAlberta.

\bibliography{mybibfile}

\begin{thebibliography}{10}
\expandafter\ifx\csname url\endcsname\relax
  \def\url#1{\texttt{#1}}\fi
\expandafter\ifx\csname urlprefix\endcsname\relax\def\urlprefix{URL }\fi
\expandafter\ifx\csname href\endcsname\relax
  \def\href#1#2{#2} \def\path#1{#1}\fi

\bibitem{Lukumon}
X.~Luo, L.~O. Oyedele, A.~O. Ajayi, O.~O. Akinade, J.~M.~D. Delgado, H.~A.
  Owolabi, A.~Ahmed, Genetic algorithm-determined deep feedforward neural
  network architecture for predicting electricity consumption in real
  buildings, Energy and AI 2 (2020) 100015.

\bibitem{SKOMSKI2020}
E.~Skomski, J.-Y. Lee, W.~Kim, V.~Chandan, S.~Katipamula, B.~Hutchinson,
  Sequence-to-sequence neural networks for short-term electrical load
  forecasting in commercial office buildings, Energy and Buildings 226 (2020)
  110350.

\bibitem{KIM2019}
Y.~Kim, H.~gu~Son, S.~Kim, Short term electricity load forecasting for
  institutional buildings, Energy Reports 5 (2019) 1270--1280.

\bibitem{Yaxing}
S.~Chen, Y.~Ren, D.~Friedrich, Z.~Yu, J.~Yu, Prediction of office building
  electricity demand using artificial neural network by splitting the time
  horizon for different occupancy rates, Energy and AI 5 (2021) 100093.

\bibitem{VANDERMEER}
D.~{van der Meer}, M.~Shepero, A.~Svensson, J.~Widén, J.~Munkhammar,
  Probabilistic forecasting of electricity consumption, photovoltaic power
  generation and net demand of an individual building using gaussian processes,
  Applied Energy 213 (2018) 195--207.

\bibitem{MUNKHAMMAR2021}
J.~Munkhammar, D.~{van der Meer}, J.~Widén, Very short term load forecasting
  of residential electricity consumption using the markov-chain mixture
  distribution ({MCM}) model, Applied Energy 282 (2021) 116180.

\bibitem{Taieb}
S.~Ben~Taieb, R.~Huser, R.~J. Hyndman, M.~G. Genton, Forecasting uncertainty in
  electricity smart meter data by boosting additive quantile regression, IEEE
  Transactions on Smart Grid 7~(5) (2016) 2448--2455.

\bibitem{forecast101}
A.~Zarnani, S.~Karimi, P.~Musilek, Quantile regression and clustering models of
  prediction intervals for weather forecasts: A comparative study, Forecasting
  1~(1) (2019) 169--188.

\bibitem{Yaslan}
Y.~Yaslan, B.~Bican, Empirical mode decomposition based denoising method with
  support vector regression for time series prediction: A case study for
  electricity load forecasting, Measurement 103 (2017) 52--61.

\bibitem{Woohyun}
W.~Kim, Y.~Han, K.~J. Kim, K.-W. Song, Electricity load forecasting using
  advanced feature selection and optimal deep learning model for the variable
  refrigerant flow systems, Energy Reports 6 (2020) 2604--2618.

\bibitem{Memarzadeh}
G.~Memarzadeh, F.~Keynia, Short-term electricity load and price forecasting by
  a new optimal {LSTM-NN} based prediction algorithm, Electric Power Systems
  Research 192 (2021) 106995.

\bibitem{Xifeng}
X.~Guo, Q.~Zhao, D.~Zheng, Y.~Ning, Y.~Gao, A short-term load forecasting model
  of multi-scale cnn-lstm hybrid neural network considering the real-time
  electricity price, Energy Reports 6 (2020) 1046--1053, 2020 The 7th
  International Conference on Power and Energy Systems Engineering.

\bibitem{Zulfiqar}
M.~Sajjad, Z.~A. Khan, A.~Ullah, T.~Hussain, W.~Ullah, M.~Y. Lee, S.~W. Baik, A
  novel {CNN-GRU}-based hybrid approach for short-term residential load
  forecasting, IEEE Access 8 (2020) 143759--143768.

\bibitem{Godoy}
J.~de~Godoy, K.~Otrel-Cass, K.~H. Toft, Transformations of trust in society: A
  systematic review of how access to big data in energy systems challenges
  scandinavian culture, Energy and AI 5 (2021) 100079.

\bibitem{Vilaplana}
G.~Mor, J.~Vilaplana, S.~Danov, J.~Cipriano, F.~Solsona, D.~Chemisana,
  Empowering, a smart big data framework for sustainable electricity suppliers,
  IEEE Access 6 (2018) 71132--71142.

\bibitem{Brendan}
H.~B. McMahan, E.~Moore, D.~Ramage, S.~Hampson, B.~A. y~Arcas,
  Communication-efficient learning of deep networks from decentralized data
  (2017).
\newblock \href {http://arxiv.org/abs/1602.05629} {\path{arXiv:1602.05629}}.

\bibitem{Gholizadeh}
N.~Gholizadeh, P.~Musilek, Distributed learning applications in power systems:
  A review of methods, gaps, and challenges, Energies 14~(12).

\bibitem{MESSAOUD2020}
S.~Messaoud, A.~Bradai, S.~H.~R. Bukhari, P.~T.~A. Quang, O.~B. Ahmed, M.~Atri,
  A survey on machine learning in internet of things: Algorithms, strategies,
  and applications, Internet of Things 12 (2020) 100314.

\bibitem{Brisimi}
T.~S. Brisimi, R.~Chen, T.~Mela, A.~Olshevsky, I.~C. Paschalidis, W.~Shi,
  Federated learning of predictive models from federated electronic health
  records, International Journal of Medical Informatics 112 (2018) 59--67.

\bibitem{Subramanya}
T.~Subramanya, R.~Riggio, Centralized and federated learning for predictive vnf
  autoscaling in multi-domain 5{G} networks and beyond, IEEE Transactions on
  Network and Service Management 18~(1) (2021) 63--78.

\bibitem{FedMed}
X.~Wu, Z.~Liang, J.~Wang, Fedmed: A federated learning framework for language
  modeling, Sensors 20~(14).

\bibitem{Railway}
G.~Hua, L.~Zhu, J.~Wu, C.~Shen, L.~Zhou, Q.~Lin, Blockchain-based federated
  learning for intelligent control in heavy haul railway, IEEE Access 8 (2020)
  176830--176839.

\bibitem{Savi}
M.~Savi, F.~Olivadese, Short-term energy consumption forecasting at the edge: A
  federated learning approach, IEEE Access 9 (2021) 95949--95969.

\bibitem{Cherkaoui}
A.~Taïk, S.~Cherkaoui, Electrical load forecasting using edge computing and
  federated learning, in: ICC 2020 - 2020 IEEE International Conference on
  Communications (ICC), 2020, pp. 1--6.

\bibitem{Choong}
Y.~L. Tun, K.~Thar, C.~M. Thwal, C.~S. Hong, Federated learning based energy
  demand prediction with clustered aggregation, in: 2021 IEEE International
  Conference on Big Data and Smart Computing (BigComp), 2021, pp. 164--167.

\bibitem{Haizhou}
H.~Liu, X.~Zhang, X.~Shen, H.~Sun, A federated learning framework for smart
  grids: Securing power traces in collaborative learning (2021).
\newblock \href {http://arxiv.org/abs/2103.11870} {\path{arXiv:2103.11870}}.

\bibitem{PAU2018}
M.~Pau, E.~Patti, L.~Barbierato, A.~Estebsari, E.~Pons, F.~Ponci, A.~Monti, A
  cloud-based smart metering infrastructure for distribution grid services and
  automation, Sustainable Energy, Grids and Networks 15 (2018) 14--25,
  technologies and Methodologies in Modern Distribution Grid Automation.

\bibitem{EMnify}
An introduction to smart meter communication,
  \url{https://www.emnify.com/en/resources/smart-meter}, accessed: October 2021
  (September 2021).

\bibitem{Muzaffar}
S.~Muzaffar, A.~Afshari, Short-term load forecasts using lstm networks, Energy
  Procedia 158 (2019) 2922--2927, innovative Solutions for Energy Transitions.

\bibitem{Kharlova20}
E.~Kharlova, D.~May, P.~Musilek, Forecasting photovoltaic power production
  using a deep learning sequence to sequence model with attention, in: 2020
  International Joint Conference on Neural Networks (IJCNN), 2020, pp. 1--7.

\bibitem{9097842}
H.~Xiao, M.~A. Sotelo, Y.~Ma, B.~Cao, Y.~Zhou, Y.~Xu, R.~Wang, Z.~Li, An
  improved lstm model for behavior recognition of intelligent vehicles, IEEE
  Access 8 (2020) 101514--101527.

\bibitem{misconvergence}
L.~Li, Y.~Fan, M.~Tse, K.-Y. Lin, A review of applications in federated
  learning, Computers \& Industrial Engineering 149 (2020) 106854.

\bibitem{Haohang}
H.~Xu, J.~Li, H.~Xiong, H.~Lu, Fedmax: Enabling a highly-efficient federated
  learning framework, in: 2020 IEEE 13th International Conference on Cloud
  Computing (CLOUD), 2020, pp. 426--434.

\bibitem{Davydenko2014}
A.~Davydenko, R.~Fildes, Measuring Forecasting Accuracy: Problems and
  Recommendations (by the Example of SKU-Level Judgmental Adjustments),
  Springer Berlin Heidelberg, Berlin, Heidelberg, 2014, pp. 43--70.

\end{thebibliography}

\end{document}